\documentstyle[11pt,psfig,apalike]{article}
\long\def\tcaption#1{%
  \rule{\textwidth}{1pt}\par\vskip 9pt\caption{#1}\par\vskip -3pt\rule{\textwidth}{.5pt}\vskip 3pt
}
\def\captionpar{\newline\hspace*{2ex}}
\newtheorem{proof}{Proof}
\newtheorem{theorem}{Theorem}

\def\aff#1#2#3{{\rm aff}_{#1}(#2,~#3)}
\def\affy#1#2{{\rm aff}(#1,~#2)}
\def\wei#1#2{{\rm weight}(#1,~#2)}
\def\facw#1#2{{\rm factor}(#1,~#2)}
\def\sim#1#2#3{{\rm sim}_{#1}(#2,~#3)}
\def\simi#1#2{{\rm sim}(#1,~#2)}

\title{Learning similarity-based word sense disambiguation \\ from sparse
data}

\author{ Yael Karov  and  Shimon Edelman \\ Dept.\ of Applied
Mathematics and Computer Science \\ The Weizmann Institute of Science
\\ Rehovot 76100, Israel \\ {\sf
  [yaelk,edelman]@wisdom.weizmann.ac.il} }

\begin{document}
\oddsidemargin 1pt
\evensidemargin 1pt
\topmargin -1in

\maketitle
\bibliographystyle{fullname}

\begin{abstract}
  We describe a method for automatic word sense disambiguation using a
  text corpus and a machine-readable dictionary (MRD). The method is
  based on word similarity and context similarity measures. Words are
  considered similar if they appear in similar contexts; contexts are
  similar if they contain similar words.  The circularity of this
  definition is resolved by an iterative, converging process, in which
  the system learns from the corpus a set of typical usages for each
  of the senses of the polysemous word listed in the MRD. A new
  instance of a polysemous word is assigned the sense associated with
  the typical usage most similar to its context.  Experiments show
  that this method performs well, and can learn even from very sparse
  training data.
\end{abstract}

\section*{Introduction}
Word Sense Disambiguation (WSD) is the problem of assigning a sense to
an ambiguous word, using its context. We assume that different senses
of a word correspond to different entries in its dictionary
definition. For example, {\tt suit} has two senses listed in a
dictionary: {\em an action in court}, and {\em suit of clothes}. Given
the sentence {\em The union's lawyers are reviewing the suit}, we
would like the system to decide automatically that {\tt suit} is used
there in its court-related sense (we assume that the part of speech of
the polysemous word is known).

In recent years, text corpora have been the main source of information
for learning automatic WSD (see, e.g., \cite{Gale++92}). A typical
corpus-based algorithm constructs a training set from all contexts of
a polysemous word {\cal W} in the corpus, and uses it to learn a
classifier that maps instances of {\cal W} (each supplied with its
context) into the senses. Because learning requires that the examples
in the training set be partitioned into the different senses, and
because sense information is not available in the corpus explicitly,
this approach depends critically on manual sense tagging --- a
laborious and time-consuming process that has to be repeated for every
word, in every language, and, more likely than not, for every topic of
discourse or source of information.

The need for tagged examples creates a problem referred to in previous
works as the {\em knowledge acquisition bottleneck}: training a
disambiguator for {\cal W} requires that the examples in the corpus be
partitioned into senses, which, in turn, requires a fully operational
disambiguator. The method we propose circumvents this problem by
automatically tagging the training set examples for {\cal W} using
other examples, that do not contain~{\cal W}, but do contain related
words extracted from its dictionary definition. For instance, in the
training set for {\tt suit}, we would use, in addition to the contexts
of {\tt suit}, all the contexts of {\tt court} and of {\tt clothes} in
the corpus, because {\tt court} and {\tt clothes} appear in the MRD
entry of {\tt suit} that defines its two senses. Note that, unlike the
contexts of {\tt suit}, which may discuss either court action or
clothing, the contexts of {\tt court} are not likely to be especially
related to clothing, and, similarly, those of {\tt clothes} will
normally have little to do with lawsuits. We will use this observation
to tag the original contexts of {\tt suit}.

Another problem that affects the corpus-based WSD methods is the {\em
sparseness of data}: these methods typically rely on the statistics of
cooccurrences of words, while many of the possible cooccurrences are
not observed even in a very large corpus \cite{ChurchMercer93}.  We
address this problem in several ways. First, instead of tallying word
statistics for the examples of each sense (which may be unreliable
when the examples are few), we collect sentence-level statistics,
representing each sentence by the set of features it contains (more on
features in section~\ref{features}).  Second, we define a similarity
measure on the feature space, which allows us to pool the statistics
of similar features. Third, in addition to the examples of the
polysemous word {\cal W} in the corpus, we learn also from the
examples of all the words in the dictionary definition of~{\cal W}. In
our experiments, this resulted in a training set that could be up to
20 times larger than the set of original examples.

The rest of this paper is organized as follows. Section~\ref{sec:sim}
describes the approach we have developed. In
section~\ref{sec:results}, we report the results of tests we have
conducted on the Treebank-2 corpus. Section~\ref{sec:disc} concludes
with a discussion of related methods and a summary. Proofs and other
details of our scheme can be found in the appendix.

\section{Similarity-based disambiguation}
\label{sec:sim}

Our aim is to have the system learn to disambiguate the appearances of
a polysemous word {\cal W} with senses ${\bf s}_1, \ldots, {\bf s}_k$,
using the appearances of {\cal W} in an untagged corpus as examples.
To avoid the need to tag the training examples manually, we augment
the training set by additional sense-related examples, which we call a
{\em feedback set}. The feedback set for sense ${\bf s}_i$ of
word~{\cal W} is the union of all contexts that contain some noun
found in the entry of ${\bf s}_i({\cal W})$ in a MRD\footnote{By $MRD$
  we mean a machine-readable dictionary or a thesaurus, or any
  combination of such knowledge sources.} (high-frequency nouns, and
nouns in the intersection of any two sense entries, as well as
examples in the intersection of two feedback sets, are discarded).
The feedback sets can be augmented, in turn, by original training-set
sentences that are closely related (in a sense defined below) to one
of the feedback set sentences; these additional examples can then
attract other original examples.

The feedback sets constitute a rich source of data that are known to
be sorted by sense. Specifically, the feedback set of ${\bf s}_i$ is
known to be more closely related to ${\bf s}_i$ than to the other
senses of the same word. We rely on this observation to tag
automatically the examples of {\cal W}, as follows. Each original
sentence containing {\cal W} is assigned the sense of its most similar
sentence in the feedback sets. Two sentences are considered to be
similar insofar as they contain similar words (they do not have to
share any word); words are considered to be similar if they appear in
similar sentences. The circularity of this definition is resolved by
an iterative, converging process, described below.

\subsection{Terminology}

A {\em context}, or {\em example} of the target word {\cal W} is any
sentence that contains~{\cal W}, and (optionally) the two adjacent
sentences in the corpus. The {\em features} of a sentence are its
nouns, verbs, and the adjectives of~{\cal W} and of the nouns from
{\cal W}'s MRD definition, all used after stemming (it is also
possible to use other types of features, such as word $n$-grams or
syntactic information; see section~\ref{features}). As the number of
features in the training data can be very large, we automatically
assign each relevant feature a weight indicating the extent to which
it is indicative of the sense (see section~\ref{sec:word-weights}).
Features that appear less than two times, and features whose weight
falls under a certain threshold are excluded. A sentence is
represented by the set of the remaining relevant features it contains.

\subsection{Computation of similarity}

Our method hinges on the possibility to compute similarity between the
original contexts of~{\cal W} and the sentences in the feedback sets.
We concentrate on similarities in the way sentences use {\cal W}, and
not in their meaning. Thus, similar words tend to appear in similar
contexts, and their textual proximity to the ambiguous word~{\cal W}
is indicative of the sense of~{\cal W}. Note that contextually similar
words do not have to be synonyms, or to belong to the same lexical
category. For example, we consider the words {\em doctor} and {\em
  health} to be similar because they frequently share contexts,
although they are far removed from each other in a typical semantic
hierarchy such as the WordNet \cite{Miller++93}. Note, further, that
because we learn similarity from the training set of~{\cal W}, and not
from the entire corpus, it tends to capture regularities with respect
to the usage of~{\cal W}, rather than abstract or general
regularities.  For example, the otherwise unrelated words {\em war}
and {\em trafficking} are similar in the contexts of the polysemous
word {\em drug} ({\em narcotic}/{\em medicine}), because the
expressions {\em drug trafficking} and {\em the war on drugs} appear
in related contexts of {\em drug}. As a result, both {\em war} and
{\em trafficking} are similar in being strongly indicative of the {\em
  narcotic} sense of {\em drug}.

Words and sentences play complementary roles in our approach: a
sentence is represented by the set of words it contains, and a word
--- by the set of sentences in which it appears. Sentences are similar
to the extent they contain similar words; words are similar to the
extent they appear in similar sentences.  Although this definition is
circular, it turns out to be of great use, if applied iteratively, as
described below.

In each iteration, we update a word similarity matrix~$M^{(w)}$, whose
rows and columns are labeled by all the words encountered in the
training set of~{\cal W}. In that matrix, the cell $M^{(w)}(i,j)$
holds a value between 0 and 1, indicating the extent to which word~$i$
is contextually similar to word~$j$.  In addition, we keep and update
a separate sentence similarity matrix~$M_i^{(s)}$ for each sense~${\bf
  s}_i$ of~{\cal W} (including a matrix~$M_0^{(s)}$ that contains the
similarities of the original examples to themselves). The rows in a
sentence matrix $M_i^{(s)}$ correspond to the original examples
of~{\cal W}, and the columns --- to the original examples of~{\cal W}
for~$i=0$, and to the feedback-set examples for sense~${\bf s}_i$,
for~$i > 0$.

\begin{figure}[htb]
  \centerline{\psfig{file=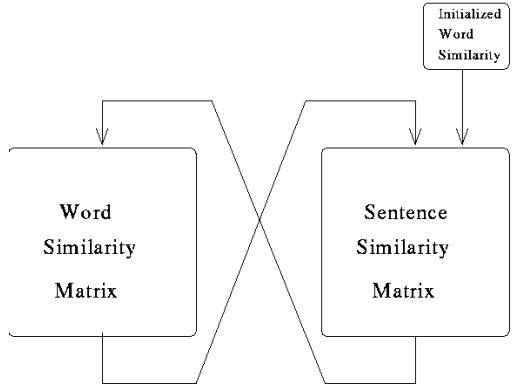,height=2in}}
\caption{ Iterative computation of word and sentence similarities. }
\label{fig:iterative}
\end{figure}

To compute the similarities, we initialize the word similarity matrix
to the identity matrix (each word is fully similar to itself, and
completely dissimilar to other words), and iterate (see
Figure~\ref{fig:iterative}):

\begin{enumerate}
\item update the sentence similarity matrices $M_i^{(s)}$, using the
  word similarity matrix $M^{(w)}$;
\item update the word similarity matrix $M^{(w)}$, using the sentence
  similarity matrices $M_i^{(s)}$.
\end{enumerate}

\noindent until the changes in the similarity values are small enough
(see section~\ref{sec:stopping} for a detailed description of the
stopping conditions; a proof of convergence appears in the appendix).

\subsubsection{The affinity formula}

The algorithm for updating the similarity matrices involves an
auxiliary relation between words and sentences, which we call {\em
  affinity}, introduced to simplify the symmetric iterative treatment
of similarity between words and sentences. A word~{\cal W} is assumed
to have a certain affinity to every sentence.  Affinity (a real number
between~$0$ and~$1$) reflects the contextual relationships
between~{\cal W} and the words of the sentence. If~{\cal W} belongs to
a sentence {\cal S}, its affinity to {\cal S} is~1; if {\cal W} is
totally unrelated to {\cal S}, the affinity is close to~0 (this is the
most common case); if {\cal W} is contextually similar to the words of
{\cal S}, its affinity to {\cal S} is between 0 and 1.  In a symmetric
manner, a sentence {\cal S} has some affinity to every word,
reflecting the similarity of {\cal S} to sentences involving that
word.

We say that a word {\em belongs} to a sentence, denoted as ${\cal W}
\in {\cal S}$, if it textually contained there; in this case, sentence
is said to {\em include} the word: ${\cal S} \ni {\cal W}$. Affinity
is then defined as follows:

\begin{eqnarray}
 \aff{n}{{\cal W}}{{\cal S}} &=& \max_{{\cal W}_i \in {\cal S}} \sim{n}{{\cal W}}{{\cal W}_i}\\
 \aff{n}{{\cal S}}{{\cal W}} &=& \max_{{\cal S}_j \ni {\cal W}}
 \sim{n}{{\cal S}}{{\cal S}_j}
\end{eqnarray}

\noindent where~$n$ denotes the iteration number.\footnote{At a first
glance it may seem that the mean rather than the maximal similarity of
{\cal W} to the words of a sentence should determine the affinity
between the two.  However, any definition of affinity that takes into
account more words than just the one with the maximal similarity
to~{\cal W}, may result in a word being directly contained in the
sentence, but having an affinity to it that is smaller than~1.} The
initial representation of a sentence, as the set of words that it
directly contains, is now augmented by a similarity-based
representation; The sentence contains more information or features
than the words directly contained in it.  Every word has some affinity
to the sentence, and the sentence can be represented by a vector
indicating the affinity of each word to it.  Similarly, every word can
be represented by the affinity of every sentence to it.  Note that
affinity is asymmetric: $\affy{\cal S}{\cal W} \neq \affy{\cal W}{\cal
  S}$, because {\cal W} may be similar to one of the words in {\cal
  S}, which, however, is not one of the topic words of {\cal S}; it is
not an important word in s.  In this case, $\affy{\cal W}{\cal S}$ is
high, because {\cal W} is similar to a word in {\cal S}, but
$\affy{\cal S}{\cal W}$ is low, because {\cal S} is not a
representative example of the usage of the word {\cal W}.

\subsubsection{The similarity formula}

We define the similarity of ${\cal W}_1$ to ${\cal W}_2$ to be the
average affinity of sentences that include ${\cal W}_1$ to those that
include ${\cal W}_2$. The similarity of a sentence ${\cal S}_1$ to
another sentence ${\cal S}_2$ is a weighted average of the affinity of
the words in ${\cal S}_1$ to those in ${\cal S}_2$:

\begin{eqnarray}
\sim{n+1}{{\cal S}_1}{{\cal S}_2} &=& 
\sum_{{\cal W} \in {\cal S}_1} \wei{{\cal W}}{{\cal S}_1}\cdot\aff{n}{{\cal
      W}}{{\cal S}_2}
\label{eq:sim-sentences}
\\
\sim{n+1}{{\cal W}_1}{{\cal W}_2} &=& \sum_{{\cal S} \ni {\cal W}_1}
\wei{{\cal S}}{{\cal W}_1}\cdot\aff{n}{{\cal S}}{{\cal W}_2}
\label{eq:sim-words}
\end{eqnarray}

\noindent where the weights sum to~1.\footnote{The weight of a word
estimates its expected contribution to the disambiguation task, and is
a product of several factors: the frequency of the word in the corpus,
its frequency in the training set relative to that in the entire
corpus; the textual distance from the target word, and its part of
speech (more details on word weights appear in
section~\protect\ref{sec:word-weights}).  All the sentences that
include a given word are assigned identical weights.}

%

\subsubsection{The importance of iteration}
\label{sec:example}

Initially, only identical words are considered similar, so that
$\affy{{\cal W}}{{\cal S}}=1$ if ${\cal W} \in {\cal S}$; the affinity
is zero otherwise.  Thus, in the first iteration, the similarity
between ${\cal S}_1$ and ${\cal S}_2$ depends on the number of words
from ${\cal S}_1$ that appear in ${\cal S}_2$, divided by the length
of ${\cal S}_2$ (note that each word may carry a different weight). In
the subsequent iterations, each word ${\cal W} \in {\cal S}_1$
contributes to the similarity of ${\cal S}_1$ to ${\cal S}_2$ a value
between~0 and~1, indicating its affinity to ${\cal S}_2$, instead of
voting either 0 (if ${\cal W} \in {\cal S}_2$) or 1 (if ${\cal W}
\not\in {\cal S}_2$).  Analogously, sentences contribute values to
word similarity.

One may view the iterations as successively capturing parameterized
``genealogical'' relationships. Let words that share contexts be
called direct relatives; then words that share neighbors (have similar
cooccurrence patterns) are {\em once-removed} relatives.  These two
family relationships are captured by the first iteration, and also by
most traditional similarity measures, which are based on
cooccurrences. The second iteration then brings together {\em
  twice-removed} relatives.  The third iteration captures higher
similarity relationships, and so on. Note that the level of
relationship here is a gradually consolidated real-valued quantity,
and is dictated by the amount and the quality of the evidence gleaned
from the corpus; it is not an all-or-none ``relatedness'' tag, as in
genealogy.

The following simple example demonstrates the difference between our
similarity measure and pure cooccurrence-based similarity measures,
which cannot capture higher-order relationships.  Consider the set of
three sentence fragments:

\begin{itemize}
\item[s1:] {\it eat banana}
\item[s2:] {\it taste banana}
\item[s3:] {\it eat apple}
\end{itemize}

In this ``corpus,'' the similarity of {\em taste} and {\em apple},
according to the cooccurrence-based methods, is~0, because the
contexts of these two words are disjoint. In comparison, our iterative
algorithm will capture some similarity:

\begin{itemize}
\item {\em Initialization.} Every word is similar to itself only.
\item {\em First iteration.} The sentences {\it eat banana} and {\it
  eat apple} have similarity of~$0.5$, because of the common word {\em
  eat}. Furthermore, the sentences {\it eat banana} and {\it taste
  banana} have similarity~$0.5$:

\begin{itemize}
\item {\em banana} is learned to be similar to {\em apple} because of
  their common usage ({\it eat banana} and {\it eat apple});
\item {\em taste} is similar to {\em eat} because of their common
  usage ({\it taste banana} and {\it eat banana});
\item {\em taste} and {\em apple} are not similar (yet).
\end{itemize}

\item {\em Second iteration.} The sentence {\it taste banana} has now
  some similarity to {\it eat apple}, because in the previous
  iteration {\em taste} was similar to {\em eat} and {\em banana} was
  similar to {\em apple}. The word {\em taste} is now similar to {\em
    apple} because the {\em taste} sentence ({\it taste banana}) is
  similar to the {\em apple} sentence ({\it eat apple}). Yet, {\em
    banana} is more similar to {\em apple} than {\em taste}, because
  the similarity value of {\em banana} and {\em apple} further
  increases in the second iteration.

\end{itemize}

\noindent This simple example demonstrates the transitivity of our
similarity measure, which allows it to extract high-order contextual
relationships.  In more complex situations, the transitivity-dependent
spread of similarity is slower, because each word is represented by
many more sentences. 
Iteration stops when the changes in the similarity values are small
enough (see section~\ref{sec:stopping}). In practice, this happens
after about three iterations, which, intuitively, suffice to exhaust
the transitive exploration of similarities.  After that, although the
similarity values may continue to increase, their rank order does not
change significantly. That is, if in the third iteration a sentence
{\cal S} was more similar to ${\cal T}_1$ than to ${\cal T}_2$, this
order will, by and large, prevail also in the subsequent iterations,
even though the similarity values may still increase. 

The most important properties of the similarity computation algorithm
are convergence (see appendix~\ref{apdx:proofs}), and utility in
supporting disambiguation (described in section~\ref{sec:results});
three other properties are as follows.  First, word similarity
computed according to the above algorithm is asymmetric.  For example,
{\em drug} is more similar to {\em traffic} than {\em traffic} is to
{\em drug}, because {\em traffic} is mentioned more frequently in {\em
  drug} contexts than {\em drug} is mentioned in contexts of {\em
  traffic} (which has many other usages).  Likewise, sentence
similarity is asymmetric: if ${\cal S}_1$ is fully contained in ${\cal
  S}_2$, then $\simi{{\cal S}_1}{{\cal S}_2} = 1$, whereas
$\simi{{\cal S}_2}{{\cal S}_1} < 1$.  Second, words with a small count
in the training set will have unreliable similarity values. These,
however, are multiplied by a very low weight when used in sentence
similarity evaluation, because the frequency in the training set is
taken into account in computing the word weights.  Third, in the
computation of $\simi{{\cal W}_1}{{\cal W}_2}$ for a very frequent
${\cal W}_2$, the set of its sentences is very large, potentially
inflating the affinity of ${\cal W}_1$ to the sentences that contain
${\cal W}_2$.  We counter this tendency by multiplying $\simi{{\cal
    W}_1}{{\cal W}_2}$ by a weight that is reciprocally related to the
global frequency of ${\cal W}_2$.

\subsection{Using similarity to tag the training set}

Following convergence, each sentence in the training set is assigned
the sense of its most similar sentence in one of the feedback sets of
sense~${\bf s}_i$, using the final sentence similarity matrix. 
Note that some sentences in the training set belong also to one of the
feedback sets, because they contain words from the MRD
definition of the target word. Those sentences are automatically
assigned the sense of the feedback set to which they belong, since they 
are most similar to themselves. Note also that an original training-set 
sentence ${\cal S}$ can be attracted to a sentence {\cal F} from a feedback 
set, even if {\cal S} and {\cal F} do not share any word, because of the 
transitivity of the similarity measure.

\subsection{Learning the typical uses of each sense}

We partition the examples of each sense into {\em typical use} sets,
by grouping all the sentences that were attracted to the same
feedback-set sentence. That sentence, and all the original sentences
attracted to it, form a class of examples for a typical usage.
Feedback-set examples that did not attract any original sentences are
discarded. If the number of resulting classes is too high, further
clustering can be carried out on the basis of the distance metric
defined by $1 - \simi{x}{y}$, where $\simi{x}{y}$ are values taken
from the final sentence similarity matrix.  

A typical usage of a sense is represented by the affinity information
generalized from its examples.  For each word {\cal W}, and each
cluster~$C$ of examples of the same usage, we define:

\begin{eqnarray}
  \affy{{\cal W}}{C} & = & \max_{{\cal S} \in C} \affy{{\cal W}}{{\cal
      S}} \\ & = & \max_{{\cal S} \in C} \max_{{\cal W}_i \in {\cal
      S}} \simi{{\cal W}}{{\cal W}_i}
\end{eqnarray}

\noindent For each cluster we construct its affinity vector, whose
$i$'th component indicates the affinity of word~$i$ to the cluster.
It suffices to generalize the affinity information (rather than
similarity), because new examples are judged on the basis of their
similarity {\bf to} each cluster: in the computation of $\simi{{\cal
    S}_1}{{\cal S}_2}$ (equation~\ref{eq:sim-sentences}), the only
information concerning ${\cal S}_2$ is its affinity values.

\subsection{Testing new examples}

Given a new sentence~{\cal S} containing a target word~{\cal W}, we
determine its sense by computing the similarity of~{\cal S} to each of
the previously obtained clusters~$C_k$, and returning the sense of the
most similar cluster:

\begin{eqnarray}
\simi{{\cal S}_{new}}{C_k} &=& \sum_{{\cal W} \in {\cal S}_{new}}
\wei{{\cal W}}{{\cal S}_{new}}\cdot\affy{{\cal W}}{{C_k}} \\
\simi{{\cal S}_{new}}{{\bf s}_i} &=& \max_{C \in {\bf s}_i}
\simi{{\cal S}_{new}}{C}
\end{eqnarray} 

\section{Experimental evaluation of the method}
\label{sec:results}

We tested the algorithm on the Treebank-2 corpus, which contains
1~million words from the Wall Street Journal, 1989, and is considered
a small corpus for the present task.  As the MRD, we used a
combination of the Webster, the Oxford and the WordNet online
dictionaries (the latter used as a thesaurus only; see
section~\ref{sec:wordnet}).  During the development and the tuning of
the algorithm, we used the method of pseudo-words
\cite{Gale++92,Schutze92}, to save the need for manual verification of
the resulting sense tags.

The final algorithm was tested on a total of 500 examples of four
polysemous words: {\em drug}, {\em sentence}, {\em suit}, and {\em
  player} (see Table~\ref{table:results}).  The relatively small
number of polysemous words we studied was dictated by the size and
nature of the corpus (we are currently testing additional words, using
texts from the British National Corpus).

\begin{table}
\tcaption{A summary of the experimental results on four polysemous words.}
\label{table:results}
\begin{center}
\begin{tabular*}{32pc}{c@{\extracolsep{\fill}}ccccc}
Word & Senses & Sample & Feedback &  \% correct &  \% correct \\
     &        & Size & Size &per sense & total \\
\hline
drug & narcotic    & 65 & 100 & 92.3  &  90.5 \\
      & medicine   & 83 & 65 &   89.1 & \\
\hline
sentence & judgement & 23 & 327 & 100  &  92.5 \\
         & grammar   & 4 & 42 &   50 & \\
\hline
suit & court    & 212 & 1461 & 98.59  &  94.8 \\
      & garment   & 21 & 81 &   55 &    \\
\hline
player & performer    & 48 & 230 & 87.5  &  92.3 \\
      & participant    & 44 & 1552 &   97.7 & \\
\hline
\end{tabular*}
\end{center}
\end{table}

The average success rate of our algorithm was $92\%$.  The original
training set (before the addition of the feedback sets) consisted of a
few dozen examples, in comparison to thousands of examples needed in
other corpus-based methods \cite{Schutze92,Yarowsky95}. 

Results on two of the words on which we tested our algorithm ({\em
drug} and {\em suit}) have been also reported in the works of Schutze
and Yarowsky. It is interesting to compare the performance of the
different methods on these words. On the word {\em drug}, our
algorithm achieved performance of $90.5\%$, after being trained on 148
examples (contexts). In comparison, \cite{Yarowsky95} achieved
$91.4\%$ correct performance, using 1380 contexts and the dictionary
definitions in training.\footnote{Yarowsky subsequently improved that
  result to $93.9\%$, using his ``one sense per discourse''
  constraint. We expect that a similar improvement could be achieved
  if that constraint were used in conjunction with our method. } On
the word {\em suit}, our method achieved performance of $94.8\%$,
using 233 training contexts; in comparison, \cite{Schutze92} achieved
$95\%$ correct performance, using 8206 contexts. In summary, our
algorithm achieved performance comparable to some of the best reported
results, using much less data for training. This feature of our
approach is important, because the size of the available training set
is usually severely constrained for most senses of most words
\cite{Gale++92}.  Finally, we note that, as in most corpus-based
methods, supplying additional examples is expected to improve the
performance.

\begin{figure}[thb]
  \centerline{\psfig{file=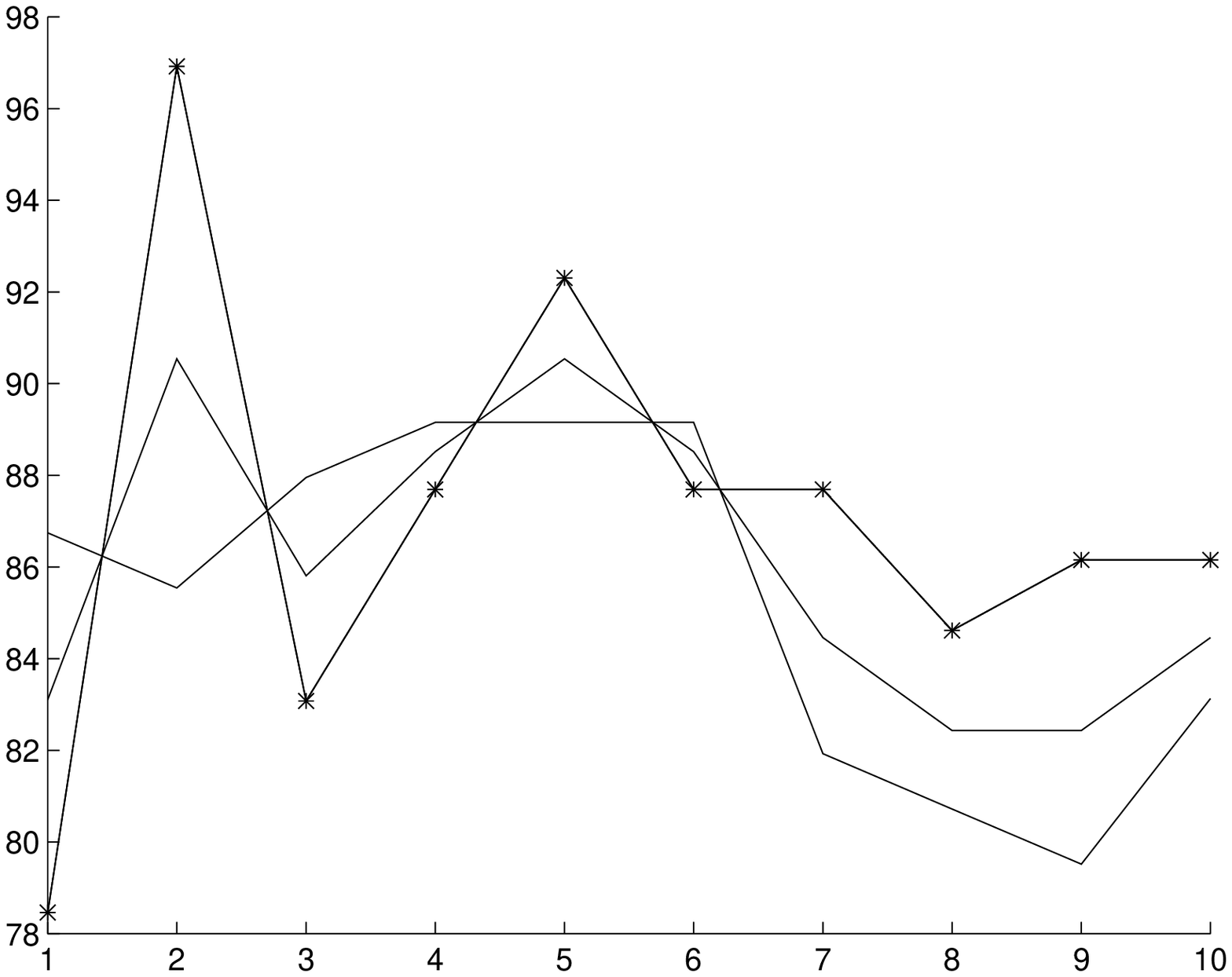,height=2in} }
\caption{The {\em drug} experiment; the change in the disambiguation
  performance with iteration number is plotted separately for each
  sense.  The asterisk marks the plot of the success rate for the {\em
    narcotic} sense. }
\label{fig:iter}
\end{figure}

We now present in detail several of the results obtained with the
word~{\em drug}. A plot of the improvement in the performance vs.\ 
iteration number appears in Figure~\ref{fig:iter}.  The success rate
is plotted for each sense, and for the weighted average of both senses
we considered (the weights are proportional to the number of examples
of each sense).

\begin{figure}[thb]
\centerline{\psfig{file=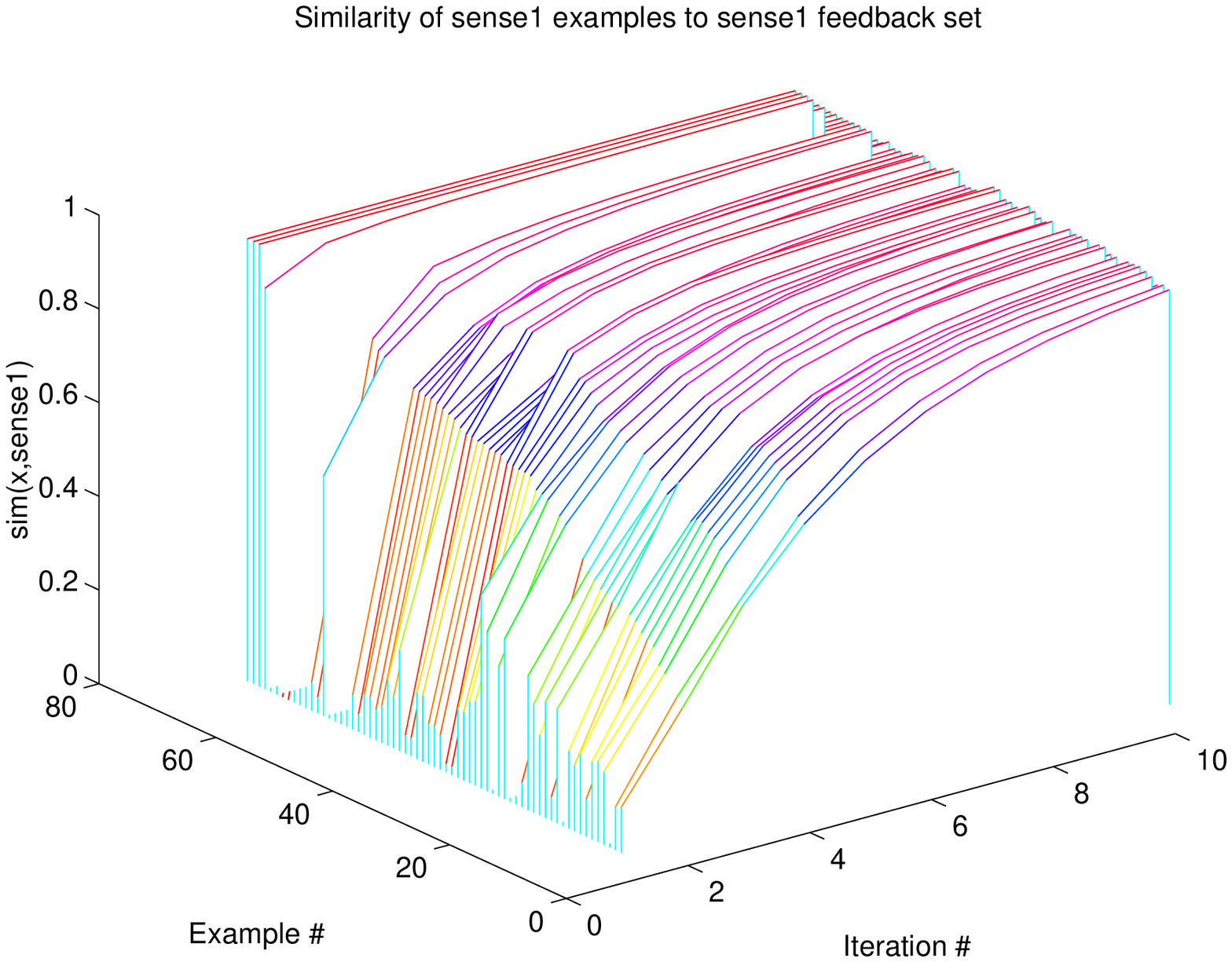,height=2in}
}
\caption{The {\em drug} experiment; example runs, sorted by the
  second-iteration similarity values.}
\label{fig:sim-increase}
\end{figure}

Figure~\ref{fig:sim-increase} shows how the similarity values develop
with iteration number. For each example {\cal S} of the {\em narcotic}
sense of {\em drug}, the value of $\sim{n}{{\cal S}}{{\em narcotic}}$
  increases with~$n$.  Note that after several iterations the
  similarity values are close to~1, and, because they are bounded
  by~1, they cannot change significantly with further iterations.

\begin{figure}[thb]
\centerline{\
\psfig{file=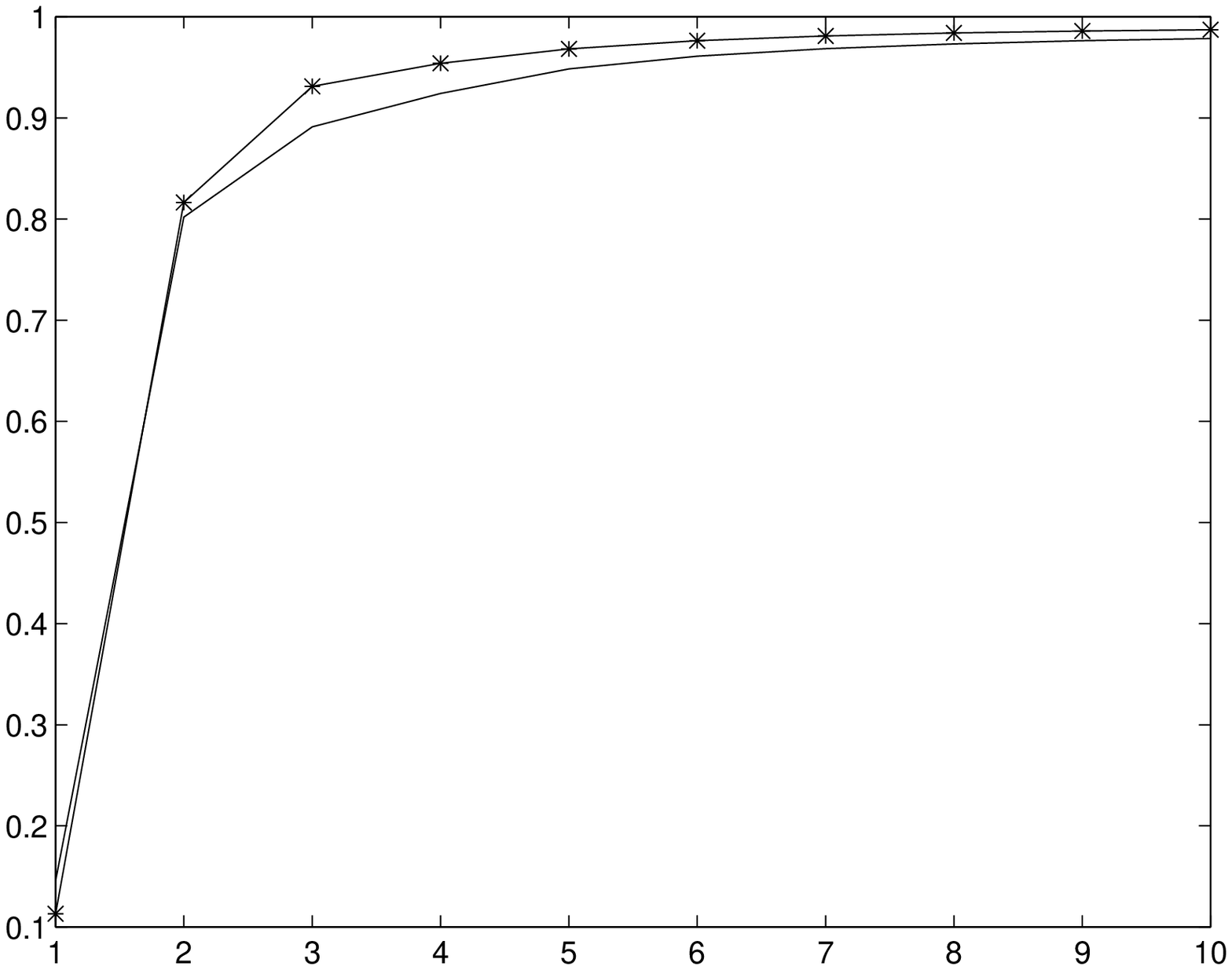,height=2in}
}
\caption{The {\em drug} experiment; the similarity between a {\em
    narcotic}-sense example to each of the two senses. The asterisk
  marks the plot for the {\em narcotic} sense. The sentence was {\em The
    American people and their government also woke up too late to the
    menace drugs posed to the moral structure of their country.}
  \captionpar 
  The word {\em menace} which is a hint for the {\em narcotic} sense
  in this sentence, did not help in the first iteration, because it
  did not appear in the {\em narcotic} feedback set at all.  Thus, in
  iteration~1, the similarity of this sentence to the {\em medicine}
  sense was~$0.15$, vs.\ similarity of~$0.1$ to the {\em narcotic}
  sense. In iteration~2, {\em menace } was learned to be similar to
  other {\em narcotic}-related words, yielding a small advantage for
  the {\em narcotic} sense. In iteration~3, further similarity values
  were updated, and there was a clear advantage to the {\em narcotic}
  sense (0.93, vs.\ 0.89 for {\em medicine}).  }
\label{fig:iterhelp}
\end{figure}

Figure~\ref{fig:iterhelp} compares the similarities of a {\em
narcotic} example to the {\em narcotic} sense and to the {\em
medicine} sense, for each iteration.  The {\em medicine} sense
assignment, made in the first iteration, has been corrected in the
following iterations.

Table~\ref{table:sim} shows the most similar words found for the words
with the highest weights in the {\em drug} example (low-similarity
words have been omitted). Note that the similarity is contextual, and
is affected by the polysemous target word. For example, {\em
  trafficking} was found to be similar to {\em crime}, because in {\em
  drug} contexts the expressions {\it drug trafficking} and {\em
  crime} are highly related. In general, {\em trafficking} and {\em
  crime} need not be similar, of course.

\begin{table}[thbp]
{\small
\begin{tabular}{ll}
Word & Most contextually similar words \\
\hline
The {$medicine$} sense: &\\
\hline
medication & antibiotic blood prescription medicine percentage pressure \\
\hline
prescription & analyst antibiotic blood campaign introduction law line-up medication medicine\\
         & percentage print profit publicity quarter sedative state television 
tranquilizer use\\
\hline
medicine & prescription campaign competition dollar earnings law  manufacturing \\
        & margin print product publicity quarter result sale saving sedative \\
        & staff state television tranquilizer unit use\\
\hline
disease & antibiotic blood line-up medication medicine prescription\\
\hline
symptom & hypoglycemia insulin warning manufacturer product \\
        & plant animal death diabetic evidence finding metabolism study\\
\hline
insulin & hypoglycemia manufacturer product symptom warning\\
        & death diabetic finding report study \\
\hline
tranquilizer & campaign law medicine prescription print publicity sedative \\
        & television use analyst profit state\\
\hline
dose & appeal death impact injury liability manufacturer miscarriage refusing ruling\\
     & diethylstilbestrol hormone damage effect female prospect state\\
\hline
The {$narcotic$} sense: & \\
\hline
consumer & distributor effort cessation consumption country reduction requirement \\
        & victory battle capacity cartel government mafia newspaper people\\
\hline
mafia & terrorism censorship dictatorship newspaper press brother nothing aspiration \\
        & assassination editor leader politics rise action country doubt freedom \\
        & mafioso medium menace solidarity structure trade world\\
\hline
terrorism & censorship doubt freedom mafia medium menace newspaper \\
        & press solidarity structure\\
\hline
murder & capital-punishment symbolism trafficking furor killing substance crime \\
        & restaurant law  bill case  problem\\
\hline
menace & terrorism freedom solidarity structure medium press censorship country doubt \\
        & mafia newspaper way attack government magnitude people relation threat world\\
\hline
trafficking & crime capital-punishment furor killing murder restaurant substance symbolism\\
\hline
dictatorship & aspiration brother editor mafia nothing politics press \\
        & assassination censorship leader newspaper rise terrorism\\
\hline
assassination & brother censorship dictatorship mafia nothing press terrorism \\
        & aspiration editor leader newspaper politics rise\\
\hline
laundering & army lot money arsenal baron economy explosive government hand \\
        & materiel military none opinion portion talk\\
\hline
censorship & mafia newspaper press terrorism country doubt freedom \\
         & medium menace solidarity structure\\
\hline
\end{tabular}
}
\caption[]{\small The {\em drug} experiment; the nearest neighbors
  of the highest-weight words.  The words in the entries are those
  with the highest weights, whose similarity values have, therefore,
  the greatest effect.  Note that the similarity is contextual, and is
  highly dependent on the polysemous target word. For example, {\em
    trafficking} was found to be similar to {\em crime}, because in
  the {\em drug} contexts the expressions {\em drug trafficking} and
  {\em crime} are highly related. In general, {\em trafficking} and
  {\em crime} need not be similar, of course.  Also note that the
  similarity is affected by the training corpus. For example, in the
  Wall Street Journal, the word {\em medicine} is mentioned mostly in
  contexts of making profit, and in advertisements.  Thus, in the {\em
    medicine} cluster there one finds words such as {\em analyst,
    campaign, profit, quarter, dollar}, which serve as hints for the
  {\it medicine} sense. Although {\em profit} and {\em medicine} are
  not closely related semantically (relative to a more balanced corpus
  than WSJ), their contexts in the WSJ contain words that are
  similarly indicative of the sense of the target word. This kind of
  similarity, therefore, suits its purpose, which is sense
  disambiguation, although it may run counter to some of our
  intuitions regarding general semantic similarity.  }
\label{table:sim}
\end{table}

\section{Discussion}
\label{sec:disc}

We now discuss in some detail the choices made at the different stages
of the development of the present method, and its relationship to some
of the previous works on word sense disambiguation.

\subsection{Flexible sense distinctions}

The possibility of strict definition of each sense of a polysemous
word, and the possibility of unambiguous assignment of a given sense
in a given situation are, in themselves, nontrivial issues in
philosophy \cite{Quine60} and linguistics \cite{Weinreich80,Cruse86}.
Different dictionaries often disagree on the definitions; the split
into senses may also depend on task at hand. Thus, it is important to
maintain the possibility of flexible distinction of the different
senses, e.g., by letting this distinction be determined by an external
knowledge source such as a thesaurus or a dictionary. Although this
requirement may seem trivial, most corpus-based methods do not, in
fact, allow such flexibility. For example, defining the senses by the
possible translations of the word
\cite{DaganItai91,Brown++91,Gale++92}, by the Roget's categories
\cite{Yarowsky92}, or by clustering \cite{Schutze92} yields a grouping
that does not always conform to the desired sense distinctions. In
comparison, our reliance on the MRD the definition of senses in the
initialization of the learning process guarantees the required
flexibility in setting the sense distinctions.  Pure MRD-based methods
allow the same flexibility, but do not yield good results, because the
definitions alone do not contain enough information for
disambiguation.

\subsection{Sentence features}
\label{features}

Different polysemous words may benefit from different types of
features of the context sentences.  Polysemous words for which
distinct senses tend to appear in different topics can be
disambiguated using single words as the context features, as we did
here. Disambiguation of other polysemous words may require taking the
sentence structure into account, using $n$-grams or syntactic
constructs as features. This additional information can be
incorporated into our method, by (1) extracting features such as
nouns, verbs, adjectives of the target word, bi-grams, tri-grams, and
subject-verb or verb-object pairs, (2) discarding features with a low
weight (cf.\ section~\ref{sec:word-weights}), and (3) using the
remaining features instead of single words (i.e., by representing a
sentence by the set of significant features it contains, and a feature
--- by the set of sentences in which it appears).

\subsection{Using WordNet}
\label{sec:wordnet}

The initialization of the word similarity matrix using WordNet (a
hand-crafted semantic network arranged in a hierarchical structure;
\cite{Miller++93}) may seem to be advantageous over simply setting it
to the identity matrix, as we have done. To compare these two
approaches, we tried to set the initial similarity between two words
to the WordNet path length between their nodes \cite{Lee++93}, and
then learn the similarity values iteratively.  This, however, led to
worse performance than the simple identity-matrix initialization.

There are several possible reasons for the poor performance of WordNet
in this comparison. First, WordNet is not designed to capture
contextual similarity. For example, in WordNet, {\em hospital} and
{\em doctor} have no common ancestor, and hence their similarity is~0,
while {\em doctor} and {\em lawyer} are quite similar, because both
designate professionals, humans, and living things. Note that,
contextually, {\em doctor} should be more similar to {\em hospital}
than to {\em lawyer}.  Second, we found that the WordNet similarity
values dominated the contextual similarity computed in the iterative
process, preventing the transitive effects of contextual similarity
from taking over.  Third, the tree distance in itself does not always
correspond to the intuitive notion of similarity, because different
concepts appear at different level of abstraction, and have a
different number of nested sub-concepts. For example, a certain
distance between two nodes may result from: (1) the nodes being
semantically close, but separated by a large distance, stemming from a
high level of detail in the related synsets, or from (2) the nodes
being semantically far from each other.\footnote{Resnik (1995)
  \nocite{Resnik95} recently suggested to overcome this particular
  difficulty by a different measure that takes into account the
  informativeness of the most specific common ancestor of the two
  words. }

\subsection{Ignoring irrelevant examples}

The feedback sets we use in training the system may contain noise, in
the form of irrelevant examples that are collected along with the
relevant and useful ones. For instance, in one of the definitions of
{\em bank} in WordNet, we find {\em bar}, which, in turn, has many
other senses that are not related to {\em bank}. Although these
unrelated senses contribute examples to the feedback set, our system
is hardly affected by this noise, because we do not collect statistics
on the feedback sets (i.e., our method is not based on mere
cooccurrence frequencies, as most other corpus-based methods are). The
relevant examples in the feedback set of the sense~${\bf s}_i$ will
attract the examples of~${\bf s}_i$; the irrelevant examples, will not
attract the examples of~${\bf s}_i$, but neither will they do damage,
because they are not expected to attract examples of~${\bf s}_j$ ($j
\neq i$).

\subsection{Related work}

\subsubsection{The knowledge acquisition bottleneck}

Brown et al.\ (1991) \nocite{Brown++91} and Gale et al.\ (1992)
\nocite{Gale++92} used the translations of the ambiguous word in a
bilingual corpus as sense tags. This does not obviate the need for
manual work, as producing bilingual corpora requires manual
translation work.\footnote{MRD's are, of course, also constructed
  manually, but, unlike bilingual corpora, these are existing
  resources, made for general use. } 

Dagan and Itai (1991) \nocite{DaganItai91} used a bilingual lexicon and 
a monolingual corpus, to save the need for translating the corpus. The problem
remains, however, that the word translations do not necessarily
overlap with the desired sense distinctions.

Schutze (1992) \nocite{Schutze92} clustered the examples in the
training set, and manually assigned each cluster a sense by observing
10-20 members of the cluster. Each sense was usually represented by
several clusters. Although this approach significantly decreased the
need for manual intervention, about a hundred examples had still to be
tagged manually for each word. Moreover, the resulting clusters did
not necessarily correspond to the desired sense distinctions.

Yarowsky (1992) \nocite{Yarowsky92} learned discriminators for each
Roget's category, saving the need to separate the training set into
senses. However, using such hand-crafted categories usually leads to a
coverage problem for specific domains, or for domains other than the
one for which the list of categories has been prepared.

Using MRDs for WSD was suggested in \cite{Lesk86}; several
researchers subsequently continued and improved this line of work
\cite{KrovetzCroft89,Guthrie++91,VeronisIde90}.  Unlike the
information in a corpus, the information in the MRD definitions is
presorted into senses. However, as noted above, the MRD definitions
alone do not contain enough information to allow reliable
disambiguation.  Recently, Yarowsky (1995) \nocite{Yarowsky95}
combined a MRD and a corpus in a bootstrapping process. In that work,
the definition words were used as initial sense indicators, tagging
automatically the target word examples containing them. These tagged
examples were then used as seed examples in the bootstrapping process.
In comparison, we suggest to combine further the corpus and the MRD by
use {\em all} the corpus examples of the MRD definition words, instead
of those words alone. This yields much more sense-presorted training
information.

\subsubsection{The problem of sparse data}

Most previous works define word similarity based on cooccurrence
information, and hence face a severe problem of sparse data. Many of
the possible cooccurrences are not observed even in a very large
corpus \cite{ChurchMercer93}. Our algorithm addresses this problem in
two ways. First, we replace the all-or-none indicator of cooccurrence
by a graded measure of contextual similarity. Our measure of
similarity is transitive, allowing two words to be considered similar
even if they are neither observed in the same sentence, nor share
neighbor words.  Second, we extend the training set by adding examples
of related words. The performance of our system compares favorably to
that of systems trained on sets larger by a factor of~100 (the results
described in section~\ref{sec:results} were obtained following
learning from several dozen examples, in comparison to thousands of
examples in other automatic methods).

Traditionally, the problem of sparse data is approached by estimating
the probability of unobserved cooccurrences using the actual
cooccurrences in the training set. This can be done by smoothing the
observed frequencies \cite{ChurchMercer93}, or by class-based methods
\cite{Brown++91,PereiraTishby92,Pereira++93,Hirschman86,Resnik92,Brill++90,Dagan++93}.
In comparison to these approaches, we use similarity information
throughout training, and not merely for estimating cooccurrence
statistics. This allows the system to learn successfully from very
sparse data.

\subsection{Summary}

We have described an approach to WSD that combines a corpus and a MRD
to generate an extensive data set for learning similarity-based
disambiguation.  Our system combines the advantages of corpus-based
approaches (large number of examples) with those of the MRD-based
approaches (data pre-sorted by senses), by using the MRD definitions
to direct the extraction of training information (in the form of
feedback sets) from the corpus.

In our system, a word is represented by the set of sentences in which
it appears. Accordingly, words are considered similar if they appear
in similar sentences, and sentences are considered similar if they
contain similar words. Applying this definition iteratively yields a
transitive measure of similarity under which two sentences may be
considered similar even if they do not share any word, and two words
may be considered as similar even if they do not share neighbor words.
Our experiments show that the resulting alternative to raw
cooccurrence-based similarity leads to better performance on very
sparse data.

\subsection*{Acknowledgments}
We thank Dan Roth for comments on a draft of this paper.


\newpage
\appendix

\section{Appendix}

\subsection{Stopping conditions of the iterative algorithm}
\label{sec:stopping}

Let $f_i$ be the increase in the similarity value in iteration~$i$:

\begin{equation}
  f_i({\cal X},{\cal Y}) = \sim{i}{{\cal X}}{{\cal Y}} -
  \sim{i-1}{{\cal X}}{{\cal Y}}
\end{equation}

\noindent where {\cal X}, {\cal Y} can be either words or sentences.
For each item {\cal X}, the algorithm stops updating its similarity
values to other items (that is, updating its row in the similarity
matrix) in the first iteration that satisfies $max_{{\cal Y}}
f_i({\cal X},{\cal Y}) \le \epsilon$, where $\epsilon > 0$ is a preset
threshold.

According to this stopping condition, the algorithm terminates after
at most $1 \over \epsilon$ iterations (otherwise, in $1 \over
\epsilon$ iterations with each $f_i > \epsilon$, we obtain
$\simi{{\cal X}}{{\cal Y}} > \epsilon \cdot {1 \over \epsilon} = 1$,
in contradiction to upper bound of~1 on the similarity values; see
section~\ref{apdx:proofs}).

We found that the best results are obtained within three iterations.
After that, the disambiguation results tend not to change
significantly, although the similarity values may continue to
increase.  Intuitively, the transitive exploration of similarities is
exhausted after three iterations.

\subsection{Proofs}
\label{apdx:proofs}

In the following, {\cal X}, {\cal Y} can be either words or sentences.

\begin{theorem}
 Similarity is bounded: $\sim{n}{{\cal X}}{{\cal Y}} \le 1$
\end{theorem}

\begin{proof}
  By induction on the number of iteration. At the first iteration,
  $\sim{0}{{\cal X}}{{\cal Y}} \le 1$, by initialization. Assume that
  the claim holds for~$n$, and prove for~$n+1$:
\begin{eqnarray*}
  sim_{n+1}(x,y) & = & \sum_{x_j \in x} weight(x_j,x) \max_{y_k \in y}
  sim_n(x_j,y_k) \\ &\le & \sum_{x_j \in x} weight(x_j,x) \cdot 1 ~~~
  \mbox{{\rm (by the induction hypothesis)}} \\ & = &
  1
\end{eqnarray*}
\end{proof}

\begin{theorem}
  Similarity is reflexive: $\forall {\cal X}, ~~ \simi{{\cal X}}{{\cal
      X}}=1 $
\end{theorem}

\begin{proof} 
  By induction on the number of iteration. $\sim{0}{{\cal X}}{{\cal
      X}} = 1$, by initialization. Assume that
  the claim holds for~$n$, and prove for~$n+1$:
\begin{eqnarray*}
  \sim{n+1}{{\cal X}}{{\cal X}} &=& \sum_{{\cal X}_i \in {\cal X}}
  \wei{{\cal X}_i}{{\cal X}} \cdot \max_{{\cal X}_j \in {\cal X}}
  \sim{n}{{\cal X}_i}{{\cal X}_j} \\ &\ge& \sum_{{\cal X}_i \in {\cal x}}
  \wei{{\cal X}_i}{{\cal X}} \cdot \sim{n}{{\cal X}_i}{{\cal X}_i} \\
  &=& \sum_{x_i \in x} \wei{{\cal X}_i}{{\cal X}} 
  \cdot 1 ~~~\mbox{{\rm (by the induction hypothesis)}} \\ &=& 1
\end{eqnarray*}

\noindent Thus, $\sim{n+1}{{\cal X}}{{\cal X}} \ge 1$. By theorem~1,
$\sim{n+1}{{\cal X}}{{\cal X}} \le 1$, so $\sim{n+1}{{\cal X}}{{\cal X}} =
1$.
\end{proof}

\begin{theorem}
   Similarity $\sim{n}{{\cal X}}{{\cal Y}}$ is a
  non-decreasing function of the number of iteration~$n$.
\end{theorem}
 
\begin{proof} 
  By induction on the number of iteration.  Consider the case of
  $n=1$: $\sim{1}{{\cal X}}{{\cal Y}} \ge \sim{0}{{\cal X}}{{\cal Y}}$
  (if $\sim{0}{{\cal X}}{{\cal Y}} = 1$, then ${\cal X}={\cal Y}$, and
  $\sim{1}{{\cal X}}{{\cal Y}} =1$ as well; else $\sim{0}{{\cal
      X}}{{\cal Y}} = 0$ and $\sim{1}{{\cal X}}{{\cal Y}} \ge 0 =
  \sim{0}{{\cal X}}{{\cal Y}}$).  Now, assume that the claim holds
  for~$n$, and prove for~$n+1$:
\begin{eqnarray*}
  \sim{n+1}{{\cal X}}{{\cal Y}} &-& \sim{n}{{\cal X}}{{\cal Y}} = \\
 &=& \sum_{{\cal X}_j \in {\cal X}} \wei{{\cal X}_j}{{\cal X}} \cdot
  \max_{{\cal Y}_k \in {\cal Y}} \sim{n}{{\cal X}_j}{{\cal Y}_k} -
  \sum_{{\cal X}_j \in {\cal X}} \wei{{\cal X}_j}{{\cal X}} \cdot
  \max_{{\cal Y}_k \in {\cal Y}} \sim{n-1}{{\cal X}_j}{{\cal Y}_k} \\
  &\ge& \sum_{{\cal X}_j \in {\cal X}} \wei{{\cal X}_j}{{\cal X}}
  \cdot \left(\max_{{\cal Y}_k \in {\cal Y}} \sim{n}{{\cal X}_j}{{\cal
      Y}_k} - \max_{{\cal Y}_k \in {\cal Y}} \sim{n-1}{{\cal
      X}_j}{{\cal Y}_k}\right) \\
&\ge& 0
\label{eq:increase}
\end{eqnarray*}

\noindent The last inequality hold because, by the induction
hypothesis,

\begin{eqnarray*}
\forall {\cal X}_j, {\cal Y}_k, ~~ \sim{n}{{\cal X}_j}{{\cal Y}_k}
&\ge& \sim{n-1}{{\cal X}_j}{{\cal Y}_k} \\
\max_{{\cal Y}_k \in {\cal Y}} \sim{n}{{\cal X}_j}{{\cal Y}_k} &\ge&
\max_{{\cal Y}_k \in {\cal Y}} \sim{n-1}{{\cal X}_j}{{\cal Y}_k} \\
\max_{{\cal Y}_k \in {\cal Y}} \sim{n}{{\cal X}_j}{{\cal Y}_k} -
\max_{{\cal Y}_k \in {\cal Y}} \sim{n-1}{{\cal X}_j}{{\cal Y}_k} &\ge& 0
\end{eqnarray*}

\noindent Thus, all the items under the sum are nonnegative, and so must be 
their weighted average. As a consequence, we may conclude that the iterative
estimation of similarity converges.
\end{proof}

\subsection{Word weights}
\label{sec:word-weights}

In our algorithm, the weight of a word estimates its expected
contribution to the disambiguation task, and the extent to which the
word is indicative in sentence similarity.  The weights do not change
with iterations. They are used to reduce the number of features to a
manageable size, and to exclude words that are expected to be given
unreliable similarity values.  The weight of a word is a product of
several factors: frequency in the corpus, the bias inherent in the
training set, distance from the target word, and part of speech label:

\begin{enumerate}
\item {\em Global frequency}. Frequent words are less informative of
  the sense and of the sentence similarity (e.g., the appearance of
  {\em this} in two different sentences does not indicate similarity
  between them, and does not indicate the sense of any target word).
  The contribution of frequency is $\max\{0,1 - {{\rm freq}(w) \over
    {\rm max5}_{{\cal X}}{\rm freq}({\cal X})}$, where ${\rm
    max5}_{{\cal X}}{\rm freq}({\cal X})$ is a function of the five
  highest frequencies in the corpus. This factor excludes only the
  most frequent words from further consideration.  As long as the
  frequencies are not very high, it does not label ${\cal W}_1$ whose
  frequency is twice that of ${\cal W}_2$ as less informative.

\item {\em Log likelihood factor}. Words that are indicative of the
  sense usually appear in the training set more than what would have
  been expected from their frequency in the general corpus. The log
  likelihood factor captures this tendency. It is computed as

\begin{equation}
 \log {\Pr\left({\cal W}_i \mid {\cal W}\right) \over \Pr\left({\cal
     W}_i\right)} \label{eq:likelihood} 
\end{equation}

\noindent where $\Pr({\cal W}_i)$ is estimated from the frequency of
{\cal W} in the entire corpus, and $\Pr({\cal W}_i \mid {\cal W})$ ---
from the frequency of ${\cal W}_i$ in the training set, given the
examples of the current ambiguous word ${\cal W}$ (cf.\ 
\cite{Gale++92}).\footnote{Because this estimate is unreliable for
  words with low frequencies in each sense set, Gale et al. (1992)
  suggested to interpolate between probabilities computed within the
  sub-corpus and probabilities computed over the entire corpus. In our
  case, the denominator is the frequency in the general corpus instead
  of the frequency in the sense examples, so it is more reliable.} To
avoid poor estimation for words with a low count in the training set,
we multiply the log likelihood by $\min\{1,{{\em count}({\cal W})
  \over 10}\}$ where {\em count}({\cal W}) is the number of
occurrences of {\cal W} in the training set.

\item {\em Part of speech}.  Each part of speech is assigned an
  initial weight (1.0 for nouns and 0.6 for verbs).

\item {\em Distance from the target word}. Context words that are far
  from the target word are less indicative than nearby ones. The
  contribution of this factor is reciprocally related to the
  normalized distance.
\end{enumerate}

The total weight of a word is the product of the above factors, each
normalized by the sum of factors of the words in the sentence:
$\wei{{\cal W}_i}{{\cal S}} = { \facw{{\cal W}_i}{{\cal S}} \over
  \sum_{{\cal W}_j \in {\cal S}} \facw{{\cal W}_j}{{\cal S}} }$, where
  $\facw{.}{.}$ is the weight before normalization.

\subsection{Other uses of context similarity}
\label{sec:thesaurus}

The similarity measure developed in the present paper can be used for
tasks other than word sense disambiguation. Here, we illustrate a
possible application to automatic construction of a thesaurus.

Following the training phase for a word {\cal X}, we have a word
similarity matrix for the words in the contexts of {\cal X}. Using
this matrix, we construct for each sense ${\bf s}_i$ of {\cal X} a set
of related words,~$R$:

\begin{enumerate}
\item Initialize $R$ to the set of words appearing in the MRD
  definition of ${\bf s}_i$;
\item Extend $R$ recursively: for each word in~$R$ added in the
  previous step, add its~$k$ nearest neighbors, using the similarity
  matrix.
\item Stop when no new words (or too few new words) are added.
\end{enumerate}

\noindent Upon termination, output for each sense ${\bf s}_i$ the set
of its contextually similar words~$R$.

\end{document}